\documentclass[11pt]{article}

\usepackage[a4paper,margin=1in]{geometry}
\usepackage{amsmath,amssymb}
\usepackage{booktabs}
\usepackage{array}
\usepackage{hyperref}
\usepackage{enumitem}
\usepackage{graphicx}
\usepackage{xcolor}
\usepackage{listings}

\usepackage[ruled,linesnumbered]{algorithm2e}

\SetKwInput{KwIn}{Input}
\SetKwInput{KwOut}{Output}
\DontPrintSemicolon
\SetAlgoNlRelativeSize{-1}
\SetAlCapFnt{\small}
\SetAlCapNameFnt{\small}
\SetKwComment{Comment}{/* }{ */}

\hypersetup{
	colorlinks=true,
	linkcolor=blue,
	citecolor=blue,
	urlcolor=blue
}

\title{Cyber-Physical Anomaly Detection in IoT-Enabled Smart Grids Using Machine Learning and Metaheuristic Feature Optimization}
\author{%
	\begin{minipage}{0.95\textwidth}
		\centering
		Adis ALIHOD\v{Z}I\'C\textsuperscript{1}, Eva TUBA\textsuperscript{2,3},  Milan TUBA\textsuperscript{4}\\[0.5em]
		\small
		\textsuperscript{1}Department of Mathematical and Computer Sciences, University of Sarajevo,\\
		Zmaja od Bosne 33--35, Sarajevo, 71000, Bosnia \& Herzegovina\\
		\textsuperscript{2}Singidunum University, Danijelova 32, Belgrade, 11000, Serbia\\
		\textsuperscript{3}Trinity University, 1 Trinity Pl, San Antonio, 78212, Texas, USA\\
		\textsuperscript{4}Sinergija University, Raje Banji\v{c}i\'ca, Bijeljina 76300, Bosnia \& Herzegovina\\[0.5em]
		\texttt{adis.alihodzic@pmf.unsa.ba},\texttt{etuba@ieee.org}, \texttt{tuba@ieee.org}
	\end{minipage}%
}
\date{}

\begin{document}
	
	\maketitle
	
\begin{abstract}Modern smart grids rely on dense measurement infrastructures, communication links, and intelligent field devices. Although this improves supervision and control, it also increases vulnerability to cyber-physical disruptions. Operators must distinguish physical incidents, such as faults or line disturbances, from malicious actions, such as false data injection or unauthorized command execution. This chapter investigates this problem using the well-known MSU/ORNL Power System Attack Dataset. The proposed method combines machine learning with genetic-algorithm-based feature selection. The objective is twofold: to classify attack and natural events accurately, and to determine whether a reduced set of physically informative PMU/IED measurements can support reliable detection. Several baseline models are evaluated, including logistic regression, RBF-SVM, XGBoost, Random Forest, and Extra Trees. The results show that tree-based ensemble models are the most effective for the considered dataset, with Extra Trees providing the strongest full-feature baseline. After feature selection, the GA + Extra Trees model reduces the clean PMU feature space from 112 attributes to an average of 27.4 attributes over five runs, while increasing macro-F1 from 0.9118 to 0.9212 and ROC-AUC from 0.9791 to 0.9837. These results indicate that many synchronized electrical measurements are redundant. A compact subset of phasor-based features can still provide accurate and interpretable anomaly detection in smart grids.
\end{abstract}
	
	
\section{Introduction}

Modern power systems are no longer limited to traditional electrical infrastructure. They now combine physical grid components with communication networks, measurement devices, and software-based control systems. This transition has been largely shaped by technologies such as smart meters, phasor measurement units (PMUs), intelligent electronic devices (IEDs), distributed generation, renewable energy sources, energy storage systems, electric vehicles, and IoT-based layers \cite{dileep2020survey,goudarzi2022iot,ntombela2023ev}. These developments have made the grid more flexible, observable, and responsive, but they have also introduced new cyber-physical security challenges \cite{achaal2024study,afrin2025cybersecurity}. The growing use of sensors and communication channels improves situational awareness and enables applications such as load forecasting, demand response, fault diagnosis, and faster system response \cite{achaal2024study,banad2025aiml}. At the same time, these technologies create additional points of vulnerability. Faulty measurements, compromised sensors, malicious control commands, or false alarms can influence decisions made at the control center \cite{achaal2024study,caleb2026fdia}. In other words, the reliability of the electrical grid is now closely connected to the security and dependability of its cyber components \cite{achaal2024study,afrin2025cybersecurity}. Artificial intelligence (AI) has become an important tool for analyzing data generated by smart grids. Machine learning and deep learning techniques are widely used for anomaly detection, intrusion detection, voltage stability assessment, fault classification, and energy management \cite{banad2025aiml,guato2024review,sarker2026review}. Among these applications, AI-based anomaly detection has gained particular attention because modern energy systems require fast and automated methods for identifying abnormal behavior \cite{guato2024review,sarker2026review}. AI-driven cybersecurity is also becoming increasingly important, since traditional passive defense mechanisms are often not sufficient against adaptive and evolving threats \cite{afrin2025cybersecurity,sahani2023mlids}. One of the most serious threats to smart grid operation is the false data injection attack \cite{liu2011fdia,caleb2026fdia}. In this type of attack, an adversary manipulates measurement data so that the control center receives information that appears valid but is actually incorrect. Such attacks are especially difficult to detect because they can bypass conventional bad-data detection techniques \cite{liu2011fdia,boyaci2022gnn}. Their impact can extend to state estimation, optimal power flow, relay operation, and energy management \cite{liu2011fdia,caleb2026fdia}. For this reason, spatial, temporal, and graph-based models are particularly useful, as power system measurements are interconnected and constrained by physical laws \cite{boyaci2022gnn,ruan2023superresolution,wang2025spatiotemporal}. Although AI-based methods show strong potential, several practical challenges remain. Many existing models use all available measurements, even when some features are redundant or provide little useful information. This leads to larger feature sets, which increase computational and communication requirements. Such overhead is especially problematic for edge devices, substations, and IoT gateways \cite{eynawi2024feature,jithish2023federated}. Another important issue is interpretability. Models with high accuracy are often difficult to explain \cite{pang2021dlad,farsi2026explainable}. This is a major concern in critical infrastructure, where operators need to understand why a certain event has been classified as an attack or as a natural disturbance \cite{farsi2026explainable}. In addition, centralized processing of smart grid data may raise privacy and security concerns, particularly when the data is collected from distributed devices and smart meters \cite{jithish2023federated,sahani2023mlids}. Because of these concerns, federated learning and edge learning are becoming increasingly relevant for future smart grid security solutions \cite{jithish2023federated,uddin2024fl}. In this chapter, we propose a compact machine learning framework for anomaly detection in IoT-enabled smart grids. Instead of relying on all available measurements, the proposed approach uses a genetic algorithm to identify a smaller and more informative subset of PMU/IED features. These selected features are then evaluated using tree-based ensemble classifiers. The main goal is to preserve strong detection performance while reducing dimensionality and improving the feasibility of deployment in practical environments. The framework is evaluated using the MSU/ORNL Power System Attack Dataset \cite{hink2014machine}. The problem is formulated as a binary classification task, where the objective is to distinguish between cyber-attacks and natural events. Several models are compared, including logistic regression, radial basis function support vector machines (RBF-SVM), XGBoost, Random Forest, and Extra Trees. Three feature configurations are analyzed: all attributes, PMU-only attributes, and PMU-only attributes without relay status flags. This comparison helps determine whether the classifiers primarily depend on electrical measurements or on log and status indicators. The main contributions of this chapter are summarized as follows:

\begin{enumerate}
	
	\item We present a cyber-physical anomaly detection framework for IoT-enabled smart grids based on PMU/IED measurements and machine learning models.
	
	\item We formulate the MSU/ORNL dataset as a binary classification problem aimed at distinguishing attacks from natural events, with particular attention given to clean PMU-based feature settings.
	
	\item We evaluate several baseline models using metrics suitable for imbalanced data, including balanced accuracy, macro-F1, and ROC-AUC.
	
	\item We propose a GA + Extra Trees feature selection approach. This method reduces the PMU feature space from 112 attributes to approximately 27, while improving macro-F1 and ROC-AUC compared with the full-feature Extra Trees baseline. The selected features are interpreted in terms of relevant power-system measurements, including voltage magnitudes, current magnitudes, phase angles, sequence components, system frequency, and apparent impedance.
	
\end{enumerate}

The rest of the chapter is organized as follows. Section 2 reviews related work on smart grid anomaly detection, including classical machine learning, deep learning, graph-based models, and explainable AI. Section 3 describes the smart grid architecture and the considered threat model. Section 4 presents the proposed GA-based feature optimization framework. Section 5 reports the experimental evaluation and discusses the obtained results. Section 6 discusses explainability, deployment, limitations, and future research directions. Section 7 concludes the chapter.

\section{Related Work}

Recent studies on anomaly detection in smart grids generally follow three main research directions. The first group of approaches relies on classical machine learning methods, such as logistic regression, decision trees, random forests, support vector machines, gradient boosting, and shallow neural networks \cite{hink2014machine,guato2024review,radoglou2018cart,farsi2026explainable}. These techniques are often attractive because they are relatively easy to interpret and do not require excessive computational resources. However, their performance usually depends on the quality of the selected features, which means that careful feature engineering is still an important part of the modeling process \cite{guato2024review,pang2021dlad}. The second research direction focuses on deep learning. Models such as CNNs, LSTMs, GRUs, temporal convolutional networks, autoencoders, and transformer-based architectures are able to learn complex nonlinear and temporal relationships directly from grid data \cite{he2017realtime,siniosoglou2021unified,zhao2020lstm,yang2021lstm_autoencoder,li2022secure_transformer,diaba2023ddos,wang2025spatiotemporal}. Hybrid deep learning models have achieved strong results in smart grid cybersecurity tasks, especially when dealing with large and complex datasets. Nevertheless, their practical use requires additional attention to computational cost, model interpretability, and robustness under real operating conditions \cite{afrin2025cybersecurity}. The third direction is based on graph learning. Since power systems have a natural graph structure, where buses, lines, and substations are physically connected, graph neural networks can use this topology to capture dependencies that are difficult to model with conventional methods. GNN-based approaches have been applied to FDIA detection, including stealth attack detection, attack localization, and scalable real-time detection in benchmark power systems \cite{boyaci2022gnn}. More recent spatio-temporal models extend this idea by combining graph convolution with temporal convolution, allowing the model to capture both spatial relationships among buses and the temporal evolution of system states \cite{wang2025spatiotemporal}. In addition, super-resolution perception-assisted graph deep learning has been proposed as a way to improve detection performance in FDIA scenarios \cite{ruan2023superresolution}. Explainability has also become an important topic in this field. In critical infrastructure, it is not enough for a model to simply classify an event as malicious or abnormal; operators also need to understand the reason behind that decision. For this reason, SHAP-based methods and other explainable AI techniques are increasingly used to identify the measurements that contribute most strongly to model predictions and to improve trust in automated detection systems \cite{farsi2026explainable,luo2025shap}. Recent reviews on smart grid cybersecurity also highlight the need for AI models that are not only accurate, but also explainable, compact, and robust enough for practical deployment \cite{afrin2025cybersecurity,vigneshwaran2026survey}.

\section{Smart Grid Architecture and Threat Model}

A smart grid can be viewed as a cyber-physical system in which the electrical and communication infrastructures operate together. The electrical layer includes generators, transmission lines, transformers, loads, energy storage units, renewable energy sources, breakers, and protection devices. The cyber layer consists of sensors, smart meters, PMUs, SCADA systems, communication networks, data concentrators, control centers, cloud services, and decision-support software. These two layers are closely connected, since measurements collected from the physical grid are transmitted through communication networks, processed by computational tools, and then used to generate control or protection actions that directly affect the power system. This structure improves monitoring, control, and energy management, but it also introduces new security risks. Incorrect measurements, delayed messages, compromised sensors, or malicious commands can affect state estimation, relay operation, power flow analysis, and control-center decisions. Therefore, smart grid security cannot be addressed only as a traditional power-system issue or only as a communication-security issue. It must be treated as a cyber-physical problem, because an event in one layer can directly influence the behavior of the other.

\begin{figure}[ht]
	\centering
	\includegraphics[width=\textwidth]{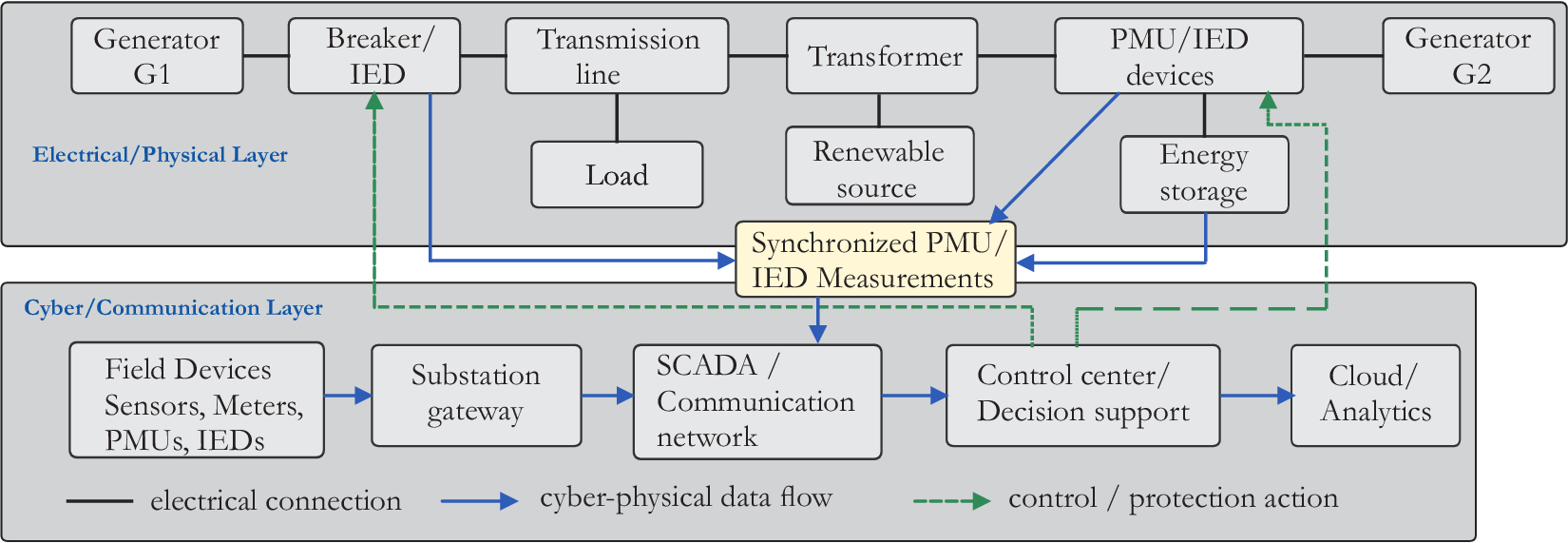}
	\caption{Simplified smart grid architecture and cyber-physical data flow.}
	\label{fig:simplified_smart_grid_architecture}
\end{figure}

\begin{figure}[ht]
	\centering
	\includegraphics[width=\textwidth]{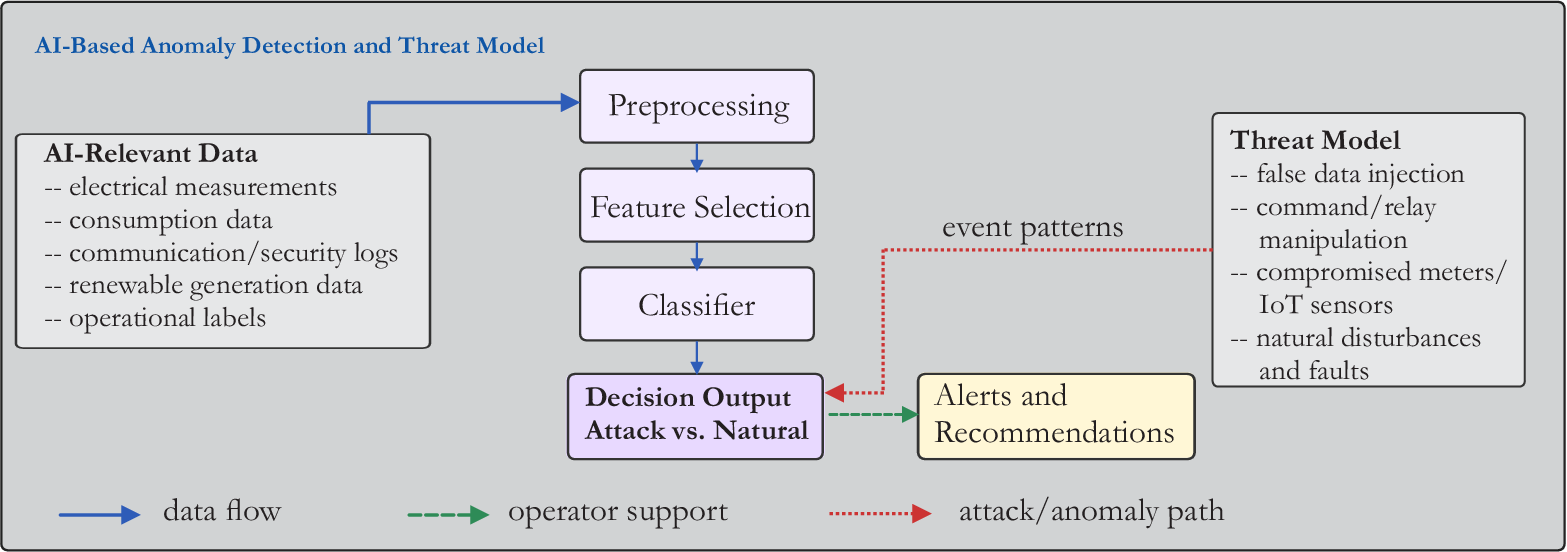}
	\caption{Simplified AI-based anomaly detection pipeline and considered threat model.}
	\label{fig:simplified_ai_threat_model}
\end{figure}

Fig.~\ref{fig:simplified_smart_grid_architecture} illustrates the simplified smart grid architecture considered in this chapter. The upper part of the figure represents the electrical and physical layer, where generators, transmission lines, transformers, loads, renewable sources, storage units, breakers, and PMU/IED devices are interconnected. The lower part represents the cyber and communication layer, where field devices, substation gateways, SCADA networks, control-center systems, and cloud analytics exchange data. The figure also indicates the direction of cyber-physical data flow from measurements toward the control center, as well as the return path of control and protection actions toward field devices and protection equipment. From the perspective of artificial intelligence, smart grid data can be divided into several main categories. Electrical measurements include bus voltages, voltage angles, active and reactive power, line currents, frequency, power factor, breaker status, PMU measurements, and relay measurements. Time-series consumption data include smart meter readings, feeder load, household-level consumption, and sub-metering values. Communication and security logs include network traffic, Snort alerts, relay logs, control-panel events, authentication records, and other indicators from the cyber layer. Renewable generation data include solar and wind production, weather-related variables, and curtailment indicators. Finally, operational labels describe the system state and may refer to normal operation, natural disturbances, cyber-attacks, attack types, fault types, or specific event scenarios. These data categories are used as input for AI-based anomaly detection, as summarized in Fig.~\ref{fig:simplified_ai_threat_model}. In such a pipeline, raw measurements and logs are first preprocessed, after which relevant features are selected and passed to a classifier. The classifier then produces a decision, for example whether the observed event is an attack or a natural disturbance. The output can be used to generate alerts and recommendations for system operators.

A cyber-physical anomaly occurs when the observed behavior of the system differs from the expected physical, temporal, or cyber behavior. Such deviations may be caused by equipment failures, natural disturbances, communication problems, measurement noise, operator actions, or malicious attacks. In practical smart grid operation, it is especially important to distinguish between a natural disturbance and a cyber-attack. A natural disturbance usually requires an electrical or operational response, while a cyber-attack requires both a power-system response and a cybersecurity response. One of the most important attack types is the false data injection attack. In this scenario, an attacker modifies selected measurements before they reach the control center. If the manipulated values are carefully designed, they may still appear physically realistic. This makes false data injection attacks particularly dangerous, because they can bypass traditional bad-data detection methods and mislead state estimation, power flow analysis, relay operation, and energy management. Recent studies indicate that graph-based and spatio-temporal models can be useful in this context, since power-system measurements are not independent but are linked through network topology and physical laws \cite{boyaci2022gnn,wang2025spatiotemporal}. Another relevant threat is command injection and relay manipulation. In this case, an attacker sends harmful commands to breakers, protection devices, or control systems. These commands may force a breaker to open when it should remain closed, or prevent it from opening during an actual fault. Relay manipulation is particularly important in distance protection systems, where an attacker may change relay settings or generate false measurements that resemble real fault conditions. If only one type of data is observed, these events may appear similar to natural disturbances. For this reason, electrical measurements and cyber-layer logs should be analyzed together whenever possible. Smart meters and IoT sensors may also be compromised. This can occur due to weak authentication, software vulnerabilities, insecure communication channels, or physical access to devices. Once compromised, a device may report false consumption values, hide energy theft, transmit misleading measurements, or participate in coordinated data manipulation. Since smart meter data can also reveal information about household behavior, privacy-preserving and distributed learning approaches are becoming increasingly important for future smart grid security systems \cite{jithish2023federated,uddin2024fl}. However, not every abnormal event in a smart grid is a cyber-attack. Natural disturbances may include short circuits, line outages, sudden load changes, generator trips, voltage instability, frequency deviations, and maintenance-related events. A reliable anomaly detection system should therefore do more than separate normal and abnormal operating states. It should also be able to distinguish malicious cyber events from physical disturbances. For this reason, binary, three-class, and multi-class classification formulations are all relevant in smart grid anomaly detection. In the experimental part of this chapter, the binary MSU/ORNL task is formulated as an \texttt{Attack} versus \texttt{Natural} classification problem. Although this is a binary setting, it provides more insight than a simple normal--abnormal formulation, because the model must distinguish malicious actions from natural power-system events. This formulation is also suitable for evaluating whether physically meaningful PMU/IED measurements are sufficient for reliable cyber-physical anomaly detection.

\section{Proposed Framework}

The proposed framework is developed for cyber-physical anomaly detection in IoT-enabled smart grids. Its purpose is not only to achieve accurate event classification, but also to reduce the number of measurements needed for reliable detection. This is particularly important because smart grid monitoring systems often include a large number of PMU/IED measurements, communication logs, relay indicators, and derived features. In practical environments, especially at the substation or edge level, using a smaller but more informative feature set can lower communication overhead, reduce memory usage, and shorten inference time. The framework is organized into several connected stages. First, data are collected from smart grid sensors, PMU/IED devices, relays, system logs, and benchmark datasets. The collected data are then preprocessed by handling missing values, replacing invalid numerical entries, normalizing features when necessary, and preparing the corresponding class labels. Depending on the structure of the available labels, the classification task may be defined as a binary, three-class, or multi-class problem. In this chapter, the experimental analysis focuses on the binary MSU/ORNL task, where the objective is to distinguish \texttt{Attack} events from \texttt{Natural} events. Let the dataset be represented as
\[
D=\{(x_i,y_i)\}_{i=1}^{N},
\]
where \(x_i \in \mathbb{R}^{d}\) denotes the feature vector of the \(i\)-th sample, while \(y_i\) represents its corresponding class label. The parameter \(d\) indicates the total number of available features. In the clean PMU-based setting used in the main experiment, both log features and relay status flags are removed. As a result, the initial feature space consists of 112 PMU/IED attributes. The goal is to learn a classifier
\[
f:\mathbb{R}^{d'} \rightarrow \mathcal{Y}, \qquad d' \leq d,
\]
where \(d'\) denotes the number of selected features and \(\mathcal{Y}\) represents the set of class labels. The selected feature subset is encoded using a binary vector
\[
z=(z_1,z_2,\ldots,z_d), \qquad z_j \in \{0,1\}.
\]
If \(z_j=1\), the \(j\)-th feature is included in the model. If \(z_j=0\), that feature is excluded. In this way, the feature selection task can be treated as a combinatorial optimization problem over binary feature masks. A useful feature subset should meet two main requirements. It should maintain strong detection performance, while also remaining compact enough for practical deployment. For this reason, the optimization criterion is designed to consider both classification quality and the number of selected features. The fitness function is defined as
\[
J(z)=\alpha\cdot\left(1-\mathrm{MacroF1}(z)\right)
+(1-\alpha)\cdot\frac{\|z\|_0}{d},
\]
where \(\mathrm{MacroF1}(z)\) denotes the validation macro-F1 score achieved by the classifier using the selected feature subset. The term \(\|z\|_0\) represents the number of selected features, while \(\alpha \in [0,1]\) controls the balance between detection performance and feature compactness. In this study, macro-F1 is used as the main optimization metric because the dataset is imbalanced, and accuracy alone may not provide a reliable view of model performance. The feature selection process is carried out using a genetic algorithm (GA) \cite{goldberg1989genetic}. Each individual in the GA population represents a binary feature mask. The initial population is generated randomly, while ensuring that every individual includes at least a minimum number of selected features. For each individual, the corresponding subset of features is used to train an evaluator classifier on the training data, and its macro-F1 score is then calculated on the validation data. After evaluation, selection, crossover, and mutation are applied to create new candidate feature subsets. Elitism is used to preserve the best-performing individuals across generations. Once the final generation is reached, the best feature mask is selected and used to train the final classifier on the full training set. The final model is then evaluated on the test set. In the experimental part of this chapter, the Extra Trees classifier \cite{geurts2006extremely} is used as the main classifier within the proposed GA-based framework. This choice is based on the baseline results, where Extra Trees achieved the strongest performance among the tested models. Random Forest is also included as an additional tree-based ensemble method for comparison. The proposed framework therefore focuses on compact and physically meaningful feature selection, rather than simply increasing model complexity. The complete procedure is presented in Algorithm~\ref{alg:ga_feature_optimization}. This framework is well suited for smart grid anomaly detection because it follows the nature of the cyber-physical problem. Instead of assuming that all measurements are equally important, it searches for a smaller subset of features that still contains enough information to distinguish between attacks and natural events. Another advantage of this approach is that the selected features can be interpreted from a power-system perspective. This makes it possible to examine whether the model relies on meaningful PMU/IED measurements, such as voltage magnitudes, current magnitudes, phase angles, sequence components, frequency, and apparent impedance.

\begin{algorithm}[ht]
	\caption{GA-Based Feature Optimization for Smart Grid Anomaly Detection}
	\label{alg:ga_feature_optimization}
	
	\KwIn{Dataset $D$, classifier $f$, number of features $d$, population size $P$, number of generations $G$, and trade-off parameter $\alpha$}
	\KwOut{Best feature subset $z^{*}$ and the final trained classifier}
	
	Preprocess the dataset by handling invalid values, imputing missing values, and preparing class labels.\;
	Split the dataset into training, validation, and test subsets.\;
	Initialize a population of binary feature masks $z_1, z_2, \ldots, z_P$.\;
	
	\ForEach{feature mask $z_p$ in the population}{
		Train the classifier $f$ using only the features selected by $z_p$.\;
		Compute the validation macro-F1 score.\;
		Compute the fitness value $J(z_p)=\alpha\cdot\left(1-\mathrm{MacroF1}(z_p)\right)
		+(1-\alpha)\cdot\frac{\|z_p\|_0}{d}$.
		
	}
	
	\For{$g=1,2,\ldots,G$}{
		Select parent masks using tournament selection.\;
		Generate new candidate masks using crossover and mutation.\;
		Evaluate the new masks using the fitness function $J$.\;
		Preserve the best individuals using elitism.\;
	}
	
	Select the best feature subset $z^{*}$.\;
	Train the final classifier using the selected features from $z^{*}$.\;
	Evaluate the final model on the test set using accuracy, balanced accuracy, F1-score, macro-F1, and ROC-AUC.\;
	\Return{$z^{*}$ and the trained classifier}\;
	
\end{algorithm}

\section{Experimental Evaluation and Discussion}

The experimental part of this chapter evaluates whether cyber attacks can be distinguished 
from natural events using compact subsets of synchronized PMU/IED measurements. The emphasis 
is not only on obtaining high classification accuracy, but also on reducing the number of 
required measurements and avoiding direct event indicators such as cyber logs or relay status 
flags.

\subsection{Dataset and Feature Sets}

The experiments use the MSU/ORNL Power System Attack Dataset. According to the dataset 
documentation, the data are generated from a cyber-physical power-system testbed consisting 
of two generators, four intelligent electronic devices (IEDs), four breakers, and two 
transmission lines. Each IED is integrated with a phasor measurement unit (PMU) and controls 
one breaker. The scenario set includes natural events, no-event operation, and attack events 
such as data injection, remote tripping command injection, and relay setting changes 
\cite{hink2014machine}. The CSV files used in this study correspond to a binary setting with two class labels: \texttt{Attack} and \texttt{Natural}. The combined dataset contains 78,377 samples. There are 55,663 attack samples and 22,714 natural-event samples. Since the dataset is imbalanced, accuracy is not sufficient as a standalone metric. For this reason, balanced accuracy, macro-F1, class-wise F1, and ROC-AUC are reported. The dataset contains 128 input attributes and one marker column. The first 116 attributes are PMU/IED measurements: four PMU/IED devices provide 29 measurements each. These  measurements include phase voltage angles, phase voltage magnitudes, phase current angles, phase current magnitudes, positive-, negative-, and zero-sequence components, frequency, frequency delta, apparent impedance, apparent impedance angle, and relay status flags. The remaining 12 attributes are control-panel logs, relay logs, and Snort alerts. Three feature sets are considered:
\begin{itemize}
	\item \textbf{all}: all 128 input features, including PMU measurements and log attributes;
	\item \textbf{pmu\_only}: 116 PMU/IED measurements, excluding control-panel, relay, and 
	Snort log attributes;
	\item \textbf{pmu\_without\_status}: 112 PMU/IED measurements, excluding both log 
	attributes and relay status flags $R1:S$, $R2:S$, $R3:S$, and $R4:S$.
\end{itemize}

The last feature set is the cleanest setting. It removes both cyber-log indicators and direct 
relay status flags. Therefore, strong performance on this feature set indicates that the 
classifier learns from physically meaningful PMU measurements rather than from direct event 
indicators.

\subsection{Preprocessing, Data Split, and Evaluation Metrics}

All CSV files are concatenated into a single dataset. Infinite values are replaced by missing 
values and then imputed using the median value computed from the training set. This is 
important because apparent impedance features may become numerically infinite when the 
corresponding current component is very small. The data are split into training, validation, and test sets using a stratified 70\%--15\%--15\% split. The validation set is used for threshold selection and GA fitness evaluation, while the test set is used only for final reporting. The main evaluation measures are accuracy, balanced accuracy, precision, recall, F1-score, macro-F1, and ROC-AUC. Because the class distribution is imbalanced, macro-F1 and balanced accuracy are treated as more informative indicators than accuracy alone.

\subsection{Baseline Models and GA-Based Feature Selection}

The following baseline models are evaluated: logistic regression, RBF-SVM, XGBoost, tuned 
XGBoost, Random Forest, and Extra Trees. Logistic regression is used as a simple linear 
baseline. RBF-SVM is included as a nonlinear kernel baseline. XGBoost is evaluated in both 
default and tuned forms. Random Forest and Extra Trees are used as tree-based bagging 
ensembles. In the experimental part of this chapter, Extra Trees \cite{geurts2006extremely} 
is used as the main classifier within the proposed GA-based framework. After the baseline comparison, a genetic algorithm is applied to the \texttt{pmu\_without\_status} feature set. Each individual in the GA population is represented by a binary mask over the 112 available features. A selected feature has value 1, while a removed feature has value 0. The fitness function combines macro-F1 and feature compactness:
\[
J(z)=\alpha(1-\mathrm{MacroF1}(z))+(1-\alpha)\frac{\|z\|_0}{d},
\]
where $z$ is the binary feature-selection vector, $\|z\|_0$ is the number of selected 
features, $d$ is the total number of available features, and $\alpha=0.95$. A smaller value 
of $J(z)$ is better. Extra Trees and Random Forest are both tested as final classifiers after 
GA-based feature selection, but Extra Trees gives the strongest results.

\subsection{Baseline and Ablation Results}

Table~\ref{tab:baseline_results} summarizes the baseline classification results. Logistic 
regression performs poorly, which indicates that the distinction between attack and natural 
events is not linearly simple. RBF-SVM improves over logistic regression in terms of the 
F1-score for the attack class, but its macro-F1 and balanced accuracy remain low. XGBoost 
improves after threshold tuning and moderate hyperparameter exploration, but it remains below 
the tree-bagging methods. Random Forest gives strong performance, while Extra Trees gives the 
best baseline result.

\begin{table}[ht]
	\centering
	\caption{Baseline classification results on the MSU/ORNL binary dataset.}
	\label{tab:baseline_results}
	\resizebox{\textwidth}{!}{%
		\begin{tabular}{llrrrrrrr}
			\toprule
			Model & Features & \#Feat. & Acc. & Bal. Acc. & Prec. & Rec. & F1 & Macro-F1 \\
			\midrule
			Logistic Regression & all & 128 & 0.5897 & 0.6396 & 0.8407 & 0.5210 & 0.6433 & 0.5802 \\
			Logistic Regression & pmu\_only & 116 & 0.5792 & 0.6381 & 0.8462 & 0.4981 & 0.6271 & 0.5722 \\
			SVM RBF & all & 128 & 0.6430 & 0.6310 & 0.8026 & 0.6596 & 0.7241 & 0.6093 \\
			SVM RBF & pmu\_only & 116 & 0.6419 & 0.6321 & 0.8041 & 0.6554 & 0.7222 & 0.6093 \\
			XGBoost & all & 128 & 0.7914 & 0.6687 & 0.7907 & 0.9606 & 0.8674 & 0.6895 \\
			XGBoost & pmu\_only & 116 & 0.7918 & 0.6683 & 0.7903 & 0.9620 & 0.8678 & 0.6891 \\
			Tuned XGBoost & all & 128 & 0.8240 & 0.7891 & -- & -- & 0.8756 & 0.7874 \\
			Tuned XGBoost & pmu\_only & 116 & 0.8278 & 0.8024 & 0.8913 & 0.8629 & 0.8768 & 0.7955 \\
			Random Forest & all & 128 & 0.9154 & 0.8701 & 0.9098 & 0.9777 & 0.9426 & 0.8909 \\
			Random Forest & pmu\_only & 116 & 0.9183 & 0.8750 & 0.9131 & 0.9780 & 0.9444 & 0.8950 \\
			Extra Trees & all & 128 & 0.9302 & 0.8985 & 0.9311 & 0.9738 & 0.9519 & 0.9121 \\
			Extra Trees & pmu\_only & 116 & 0.9312 & 0.8997 & 0.9317 & 0.9746 & 0.9526 & 0.9134 \\
			\bottomrule
		\end{tabular}%
	}
\end{table}

The results show that tree-based bagging ensembles are particularly effective for this 
dataset. Extra Trees achieves the best baseline performance with a macro-F1 score of 0.9134 
using only PMU measurements. This is important because removing cyber-log features does not 
degrade performance. On the contrary, the \texttt{pmu\_only} setting slightly improves 
macro-F1 compared with the full feature set. To verify whether the strong performance depends on direct event indicators, an ablation study is performed. First, all log attributes are removed, producing the \texttt{pmu\_only} feature set. Second, relay status flags are also removed, producing the \texttt{pmu\_without\_status} feature set. The results are shown in 
Table~\ref{tab:ablation_results}. The ablation results show that neither log attributes nor relay status flags are necessary for strong classification performance. Extra Trees remains highly effective even when both types of potentially direct indicators are removed. This supports the conclusion that the classifier relies primarily on physically meaningful PMU measurements such as voltage magnitudes, current magnitudes, phase angles, sequence components, frequency, and apparent impedance.

\begin{table}[ht]
	\centering
	\caption{Ablation study for Random Forest, XGBoost, and Extra Trees.}
	\label{tab:ablation_results}
	\resizebox{\textwidth}{!}{%
		\begin{tabular}{llrrrrr}
			\toprule
			Model & Features & \#Feat. & Acc. & Bal. Acc. & F1 & Macro-F1 \\
			\midrule
			Random Forest & all & 128 & 0.9154 & 0.8701 & 0.9426 & 0.8909 \\
			Random Forest & pmu\_only & 116 & 0.9183 & 0.8750 & 0.9444 & 0.8950 \\
			Random Forest & pmu\_without\_status & 112 & 0.9189 & 0.8762 & 0.9448 & 0.8959 \\
			Tuned XGBoost & pmu\_only & 116 & 0.8278 & 0.8024 & 0.8768 & 0.7955 \\
			Tuned XGBoost & pmu\_without\_status & 112 & 0.8289 & 0.7921 & 0.8795 & 0.7921 \\
			Extra Trees & all & 128 & 0.9302 & 0.8985 & 0.9519 & 0.9121 \\
			Extra Trees & pmu\_only & 116 & 0.9312 & 0.8997 & 0.9526 & 0.9134 \\
			Extra Trees & pmu\_without\_status & 112 & 0.9298 & 0.8984 & 0.9517 & 0.9118 \\
			\bottomrule
		\end{tabular}%
	}
\end{table}

\subsection{GA-Based Feature Selection Results}

After establishing Extra Trees as the strongest baseline, a genetic algorithm (GA) is applied to the clean \texttt{pmu\_without\_status} feature set. This setting contains 112 PMU/IED features after removing all log attributes and relay status flags. The GA is executed with five different random seeds. Table~\ref{tab:ga_extra_trees_results} reports the final Extra Trees 
performance using the GA-selected feature subsets.

\begin{table}[ht]
	\centering
	\caption{GA + Extra Trees results across five independent runs.}
	\label{tab:ga_extra_trees_results}
	\begin{tabular}{rrrrrrr}
		\toprule
		Seed & \#Selected & Acc. & Bal. Acc. & F1 & Macro-F1 & ROC-AUC \\
		\midrule
		1 & 26 & 0.9355 & 0.9073 & 0.9555 & 0.9193 & 0.9824 \\
		2 & 30 & 0.9375 & 0.9088 & 0.9569 & 0.9216 & 0.9839 \\
		3 & 25 & 0.9354 & 0.9057 & 0.9555 & 0.9189 & 0.9832 \\
		4 & 28 & 0.9391 & 0.9105 & 0.9580 & 0.9236 & 0.9847 \\
		5 & 28 & 0.9382 & 0.9103 & 0.9573 & 0.9226 & 0.9842 \\
		\midrule
		Mean & 27.4 & 0.9371 & 0.9085 & 0.9566 & 0.9212 & 0.9837 \\
		Std. & 1.95 & 0.0016 & 0.0020 & 0.0011 & 0.0020 & 0.0009 \\
		\bottomrule
	\end{tabular}
\end{table}

The GA results are stable across independent runs. The selected number of features ranges 
from 25 to 30, with an average of 27.4 selected features. This corresponds to a reduction 
from 112 to approximately 27 features, or about 75.5\% fewer input features. Despite this 
substantial reduction, the average test macro-F1 score increases to 0.9212 and the average 
ROC-AUC increases to 0.9837. Therefore, GA-based feature selection not only reduces 
dimensionality, but also improves generalization.

\subsection{Comparison with Full-Feature Extra Trees}

Table~\ref{tab:full_vs_ga} compares the best full-feature baseline with the proposed 
GA + Extra Trees approach.

\begin{table}[ht]
	\centering
	\caption{Comparison between full-feature Extra Trees and GA-selected Extra Trees.}
	\label{tab:full_vs_ga}
	\begin{tabular}{lrrrrr}
		\toprule
		Method & \#Features & Acc. & Bal. Acc. & Macro-F1 & ROC-AUC \\
		\midrule
		Extra Trees, pmu\_without\_status & 112 & 0.9298 & 0.8984 & 0.9118 & 0.9791 \\
		GA + Extra Trees, mean over 5 runs & 27.4 & 0.9371 & 0.9085 & 0.9212 & 0.9837 \\
		\bottomrule
	\end{tabular}
\end{table}

The proposed GA + Extra Trees framework improves all major metrics while using only about 
one quarter of the original clean PMU feature space. The improvement in macro-F1 is 
particularly important because the dataset is imbalanced. The improvement in balanced 
accuracy also confirms that the selected compact feature subset improves the discrimination 
of both attack and natural-event classes.

\subsection{Interpretation and Discussion}

The GA-selected subsets contain features from all four PMU/IED locations. The selected 
attributes include phase voltage magnitudes, phase current magnitudes, phase angles, 
sequence-component measurements, frequency-related features, and apparent impedance features. 
This is consistent with the physical structure of the problem. Natural events such as 
short-circuit faults and maintenance conditions affect voltages, currents, phase 
relationships, and impedance seen by the relays. Cyber attacks such as data injection, 
command injection, or relay setting manipulation may also create inconsistencies in these 
measurements. The most important point is that the selected subsets do not include control-panel logs, relay logs, Snort logs, or relay status flags. Therefore, the obtained results are not based on direct cyber-log indicators or relay-state shortcuts. Instead, the model achieves strong performance using compact sets of phasor-based electrical measurements. The experimental results support three main conclusions. First, tree-based bagging ensembles 
are more suitable for the considered MSU/ORNL binary classification task than logistic 
regression, RBF-SVM, or the tested XGBoost configurations. Second, PMU/IED measurements alone 
are sufficient for accurate discrimination between attack and natural events. Removing log 
attributes and relay status flags does not degrade the best-performing models. Third, 
GA-based feature selection provides a compact and more effective representation of the 
power-system state. From a practical smart-grid perspective, this result is important. A compact detector that uses approximately 25--30 PMU measurements instead of 112 measurements may reduce communication overhead, memory requirements, feature extraction cost, and inference 
complexity. Such a model is more suitable for edge gateways, substation-level monitoring, and 
IoT-enabled energy infrastructures. At the same time, the selected features remain physically 
meaningful and can be interpreted in terms of voltage, current, phase, frequency, and 
impedance behavior. Overall, the proposed GA + Extra Trees framework provides a strong trade-off between detection quality, compactness, and interpretability. It supports the development of secure and intelligent energy systems in which cyber-physical anomalies can be detected using a reduced but informative subset of synchronized electrical measurements.
	
\section{Explainability, Deployment, and Future Directions}

The proposed framework is not intended only as an offline classification model, but also as 
a step toward practical cyber-physical anomaly detection in next-generation smart grids. 
For this reason, three aspects are important after the experimental evaluation: operator 
trust, deployability, and the limitations that should guide future research.

\subsection{Explainable AI for Operator Trust}

A smart grid anomaly detector should support operator decision-making. If a model only reports that an event is malicious, the operator still needs to understand which measurements were responsible for the decision. Explainable AI methods such as SHAP can be used to rank features and visualize local explanations. This is particularly useful for distinguishing a physical disturbance from a cyber-attack. For example, a physical fault may produce consistent changes in voltages, currents, and relay states, while a cyber-attack may create inconsistencies between electrical measurements and communication logs. Explainability should be evaluated at two levels:
\begin{itemize}
    \item \textbf{Global explanations:} identifying which features are most important across the whole dataset;
    \item \textbf{Local explanations:} explaining why a particular event was classified as an attack, disturbance, or normal operation.
\end{itemize}

The combination of metaheuristic feature selection and explainable AI is especially attractive. Feature selection produces a compact set of measurements, while XAI explains how those measurements influence the final decision. In this way, the model becomes more transparent and more suitable for use in critical infrastructure, where operators must understand not only the classification result but also its technical justification.

\subsection{Edge and Federated Learning Deployment}

In a centralized architecture, all measurements are sent to a control center. This simplifies 
training and inference, but creates communication overhead, latency, and privacy risks. In an 
edge architecture, local devices or substation gateways perform preliminary detection. Only 
alarms, compressed features, or model updates are sent to the central operator. This approach 
can reduce latency and improve resilience. Federated learning is a natural extension for smart metering and distributed energy systems. Instead of sending raw consumption or sensor data to a central server, each local node trains a model on its own data and shares only model updates. This can help preserve privacy and reduce the exposure of household-level energy patterns \cite{jithish2023federated,uddin2024fl}. However, federated learning also introduces challenges such as non-IID data, communication cost, poisoned updates, and the need for secure aggregation. A practical deployment scenario for the proposed framework can be described as follows:
\begin{enumerate}
	\item local meters, PMUs, and IEDs collect measurements;
	\item an edge gateway computes a selected subset of features;
	\item a lightweight anomaly detector runs locally;
	\item suspicious events are transmitted to the control center;
	\item the control center applies a stronger model and XAI module;
	\item confirmed events trigger cybersecurity and electrical response procedures.
\end{enumerate}

This deployment model is consistent with the experimental findings of this chapter. Since the 
GA-selected model uses only a compact subset of PMU/IED measurements, it is more suitable for 
edge gateways, substations, and IoT-enabled energy infrastructures than a model that requires 
all available measurements.

\subsection{Practical Discussion}

The proposed approach is aligned with the needs of next-generation energy systems. It combines 
security, intelligence, and practical deployability in one framework. Security is addressed 
through cyber-physical anomaly detection. Intelligence is achieved through machine learning and 
metaheuristic feature selection. Deployability is improved by reducing the number of measurements required by the detector. The most important experimental finding is that a compact subset of PMU/IED measurements can outperform the full clean feature set. This is relevant because practical smart-grid devices, substation computers, and edge gateways may have limited computational, communication, and memory resources. A detector that uses approximately one quarter of the original PMU measurements is more attractive for deployment than a model requiring all available synchronized measurements. Another important point is that electrical knowledge should guide AI design. Smart grid data are not ordinary tabular data. Voltages, currents, phase angles, sequence components, frequency, and apparent impedance are constrained by the physical behavior of the grid. The ablation study confirms that the best models do not require cyber-log attributes or relay status flags. This reduces the risk that the classifier exploits direct event indicators instead of learning physically meaningful patterns. Overall, the proposed GA + Extra Trees framework provides a strong trade-off between detection 
quality, compactness, and interpretability. It supports the development of secure and intelligent energy systems in which cyber-physical anomalies can be detected using a reduced but informative subset of synchronized electrical measurements.

\subsection{Limitations and Future Work}

The main limitation of public benchmark datasets is that many of them are simulated or laboratory-based. Real smart grid data are difficult to obtain because of security, privacy, and regulatory constraints. Therefore, results obtained on public datasets should be interpreted carefully. Another limitation is that cyber-attacks evolve over time, while supervised models are often trained on fixed attack types. Future work should therefore include continual learning, domain adaptation, adversarial robustness, and transfer learning between simulated and real systems. Future research directions include:
\begin{itemize}
    \item graph neural networks with explicit power-system topology;
    \item physics-informed loss functions for anomaly detection;
    \item federated learning for privacy-preserving smart meter security;
    \item adversarial training against adaptive attackers;
    \item online learning for evolving cyber-physical threats;
    \item deployment of compact models on edge devices and IoT gateways;
    \item integration of anomaly detection with optimal power flow and demand response.
\end{itemize}

\section{Conclusion}
	
This chapter presented an AI-based framework for cyber-physical anomaly detection in IoT-enabled smart grids, with emphasis on compact and physically meaningful feature selection. The experimental evaluation was conducted on the MSU/ORNL Power System Attack Dataset. Several baseline models were evaluated, including logistic regression, RBF-SVM, XGBoost, tuned XGBoost, Random Forest, and Extra Trees. The results show that tree-based bagging ensembles are much more effective than the linear, kernel-based, and tested boosting baselines for the considered binary attack--natural event classification task. The strongest result was obtained by combining genetic-algorithm-based feature selection with Extra Trees. Using the clean PMU feature set without log attributes and relay status flags, the proposed GA + Extra Trees framework reduced the feature space from 112 measurements to an average of 27.4 measurements across five independent runs. This corresponds to a reduction of approximately 75.5\%. Despite this reduction, the approach improved the average test macro-F1 score to 0.9212 and the average ROC-AUC to 0.9837, outperforming the full-feature Extra Trees baseline. These findings indicate that many synchronized electrical measurements are redundant for the considered attack--natural event discrimination task. A compact subset of phasor-based measurements can provide strong detection performance while reducing communication overhead, feature extraction cost, and inference complexity. This makes the proposed approach suitable for future secure, intelligent, and edge-deployable energy infrastructures. Future work should extend the framework to three-class and multi-class settings, graph-based models with explicit power-system topology, online learning, adversarial robustness, and federated learning for privacy-preserving smart-grid security.

\end{document}